\documentclass[10pt,journal]{IEEEtran}

\usepackage{amsmath}
\usepackage{pdfpages}

\usepackage{url}
\ifCLASSOPTIONcompsoc
  \usepackage[nocompress]{cite}
\else
  \usepackage{cite}
\fi
\ifCLASSINFOpdf
\else
\fi
\usepackage{csquotes}
 \usepackage{array,multirow,graphicx}
 \usepackage{float}
\usepackage{array, makecell} %
\usepackage{tabularx,booktabs}

\begin{document}
%
\title{Unveiling the Two-Faced Truth: Disentangling Morphed Identities for Face Morphing Detection}
%
%
%

\author{Eduarda Caldeira\**, Pedro C. Neto\**, Tiago Gonçalves\**, Naser Damer, Ana F. Sequeira, Jaime S. Cardoso
\IEEEcompsocitemizethanks{\IEEEcompsocthanksitem \** These authors contributed equally.\protect\\
\IEEEcompsocthanksitem Eduarda Caldeira, Pedro C. Neto, Tiago Gonçalves, Ana F. Sequeira and Jaime Cardoso are with INESC TEC and University of Porto.\protect\\
\IEEEcompsocthanksitem Naser Damer is with the Fraunhofer Institute for Computer Graphics Research IGD and the TU Darmstadt.}
\thanks{}}

%
%

\markboth{}%
{Shell \MakeLowercase{\textit{et al.}}: Bare Advanced Demo of IEEEtran.cls for IEEE Computer Society Journals}
%



\IEEEtitleabstractindextext{%
\begin{abstract}
Morphing attacks keep threatening biometric systems, especially face recognition systems. Over time they have become simpler to perform and more realistic, as such, the usage of deep learning systems to detect these attacks has grown. At the same time, there is a constant concern regarding the lack of interpretability of deep learning models. Balancing performance and interpretability has been a difficult task for scientists. However, by leveraging domain information and proving some constraints, we have been able to develop IDistill, an interpretable method with state-of-the-art performance that provides information on both the identity separation on morph samples and their contribution to the final prediction. The domain information is learnt by an autoencoder and distilled to a classifier system in order to teach it to separate identity information. When compared to other methods in the literature it outperforms them in three out of five databases and is competitive in the remaining. 
\end{abstract}

}

\maketitle

\IEEEdisplaynontitleabstractindextext

%
\IEEEpeerreviewmaketitle

\ifCLASSOPTIONcompsoc
\IEEEraisesectionheading{\section{Introduction}\label{sec:introduction}}
\else
\section{Introduction}
\label{sec:introduction}
\fi

\IEEEPARstart{F}{ace} recognition (FR) systems have had large-scale adoption in the most diverse scenarios~\cite{wang2021deep,neto2022beyond,neto2021focusface}. Deep learning (DL) techniques have taken this and other biometric recognition systems towards above-human performance. While it also benefited biometric systems adoption, DL methods led to two problems. First, the approaches that improve the recognition power of these systems are the same to be used to design novel and dangerous attacks~\cite{damer2018morgan}. Some attacks can take the form of adversarial noise addition or be developed with FR systems in mind. The latter comprises both morphing~\cite{medvedev2022mordeephy} and presentation attacks~\cite{neto2022myope}. Besides the attacks, deep learning methods are notorious for their black-box behaviour, which compromises the understanding of both the inner workings of the model and the reasoning behind a decision. Furthermore, FR and face attack detection systems are consistently designed using problem-agnostic tools, which do not leverage domain knowledge. For a wider adoption and to be able to deploy these systems on critical scenarios, it is necessary to guarantee that their reasoning process is, at least to some extent, explained. One can explain a decision using a post-hoc approach~\cite{neto2022pic}, or directly interfering with the training behaviour of the model as stated by Neto~\textit{et al.}~\cite{neto2022explainable}. 

Face morphing attacks, which merge two images from different identities into a single image capable of misdirecting the recognition system, have progressed significantly. In other words, this attack aims to increase the number of false positives of the FR system, granting access to two distinct users. When left undetected, the fusion of two images might allow two different people to pass border control with the same passport, for example. Due to this threat, researchers focused their attention on the development of robust morphing attack detection (MAD) systems~\cite{damer2018morgan,damer2021pw,huber2022syn}. Usually, they are designed to detect if the input image is an attack or a \textit{bonafide} sample and do not include any information regarding the fused identities in their training. OrthoMAD~\cite{neto2022orthomad} aims to learn this information in an unsupervised manner by separating identity information into two orthogonal latent vectors. However, it lacks guarantees regarding the relation between the disentangled information and identity information. The recent blossoming of synthetic data generation methods, such as generative adversarial networks (GAN) and diffusion models, led to the creation of synthetic datasets with a diverse number of identities represented~\cite{damer2022privacy,damer2023mordiff}. Although these identities are usually represented only once, it suffices to increase identity diversity.

The work presented in this document builds on top of OrthoMAD premises that it is possible to disentangle information regarding different identities. The first addition is an auto-encoder model trained on the \textit{bonafide} samples to minimize the reconstruction error. The latent vector produced by the encoder is considered to be the prior of the identity information that should be present in the disentangled vectors. We further relax the orthogonality constraint to ensure that the angle between the two identity vectors, in the case of an attack, approximates the angle of the priors of their identities. To achieve this, we leverage the latent vectors of the auto-encoder for both images (before being morphed) and a knowledge distillation strategy. Finally, to further approximate the latent space of both identity vectors we replace the concatenation and classification process with a shared linear layer to be used on both vectors separately. The two predicted scores are fused afterwards. 

The main contributions of this work are the following: 1) an unexplored knowledge distillation approach based on the angle of two vectors that represent identity priors; 2) the improvement on the usage of the diverse identity set to regularize the latent spaces and the identity disentanglement; 3) a novel method designed specifically for this domain and with increased transparency regarding its inner workings; 4) an empirical validation and comparison with similar (state-of-the-art) approaches. 

This document is divided into five main sections. Besides this introductory section, the following sections include a description of the methodology, an introduction to the databases used for training and evaluation, the experimental setup designed for the experiments, the discussion of the results, and finally the conclusion. The code related to this paper is publicly available in a GitHub repository\footnote{\url{https://github.com/NetoPedro/IDistill}}.

\section{Methodology}

\begin{figure*}
    \centering
    \includegraphics[width=\linewidth]{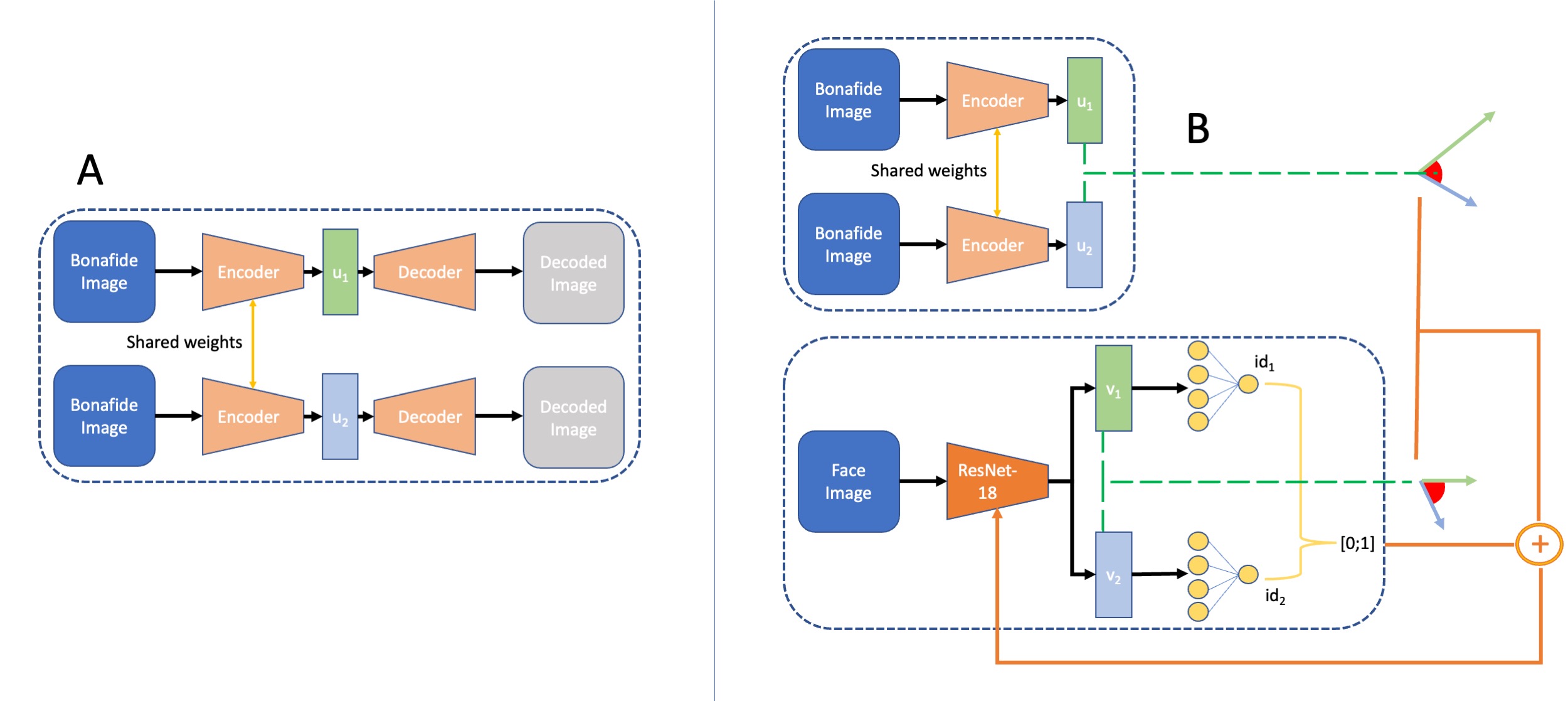}
    \caption{Overall scheme of the architecture. Part A represents the training of the autoencoder, whereas part B represents the training of the classification system. The orange line represents the backpropagation, the green one represents the calculation of the angle between vectors and the yellow represents the fusion of the two scores. Best viewed in color. }
    \label{fig:architecture}
\end{figure*}

Morphing attacks occur when two distinct identities are fused together, resulting in an image that can trick a face recognition system by containing enough information about both identities. To analyse whether information from two distinct identities is present in an image, we designed a regularisation term based in knowledge distillation (KD). As such, we call IDistill to our proposed method. The overall scheme and architecture of our proposed model is represented on Figure~\ref{fig:architecture}.

We start by training an autoencoder to reconstruct \textit{bonafide} images. This autoencoder is responsible for creating a minimalist representation of a face $I$. Alternatively to the usage of the autoencoder, we could have leveraged a pretrained face recognition system. The decision to follow with the autoencoder yields three reasons: 1) Fang~\textit{et al.}~\cite{fang2022unsupervised} has shown a difference in the reconstruction performance from abnormal and normal face images; 2) Besides being large (512-d), the latent vector face recognition systems might not contain all the information necessary for the reconstruction. 3) Encoder-Decoder have been explored for face de-morphing~\cite{banerjee2022facial}. The proposed autoencoder is based on the U-Net architecture~\cite{ronneberger2015u} and the size of the latent representation of the image was chosen as $128$. As in other reconstruction tasks the autoencoder receives the image $I$, creates a latent representation $u$ of that image using the encoder, and reconstruct it as $\tilde I$ using the decoder network.  This approach uses a mean squared error (MSE) loss function (see Eq.~\ref{eq:mse_loss}).

\begin{equation}\label{eq:mse_loss}
    L_{auto} = \sum_{i,j} (I_{ij} - \tilde I_{ij})^2
\end{equation}

The architecture of the morphing classifier is based on a ResNet-18~\cite{he2016deep} where the last fully-connected layer is replaced with two fully-connected layers that output two vectors $v$ of size 128 each. Afterwards, a fully-connected layer is used to infer if the vectors contain information of an identity or not, with each vector producing a score ($id$). The same layer is used for both vectors individually.

\begin{equation}
    id = \frac{1}{1+e^{-W^{T}v}}:
\end{equation}

Considering that the produced score holds information regarding the presence of encoded identity information on a vector $v$, the final prediction for an image $I$ is designed as follows. Given $I$, the backbone architecture produces $v_1$ and $v_2$, which will result in the identity probabilities $id_1$ and $id_2$, respectively. The probability of $I$ containing information of two distinct identities is given by $id_1*id_2$. Consequently, the \textit{bonafide} presentation score, $\tilde y$ is given by:

\begin{equation}
    \tilde y = 1-id_1*id_2
\end{equation}

For the classification task, we have introduced the Binary Cross-Entropy, $L_{BCE}$ (Equation~\ref{eq:bce}) at the level of the final fused prediction $\tilde y$. 

\begin{equation}
\label{eq:bce}
    L_{BCE} = -(y\:log(\tilde y)+(1-y)\:log(1-\tilde y))
\end{equation}

To ensure that the latent vectors $v_1$ and $v_2$ extract identity information and are aligned with the information learnt by the autoencoder, we introduce a knowledge distillation term. For attacks, this term aims to extract vectors from the morphed image that have an angle between them equal to the angle produced by the latent vectors $u_1$ and $u_2$ extracted with the encoder from the two images that originated the morphed image (Eq.~\ref{eq:attack}). We are then promoting a proximity between the autoencoder latent space and the morphing classifier latent space while handling attacks. Furthermore, we only consider the angle formed by these vectors, since their identity intensity might be diminished in the morphing process. For \textit{bonafide}, we expect one vector to hold identity information, while the other does not. As such, we designed a term that first selects the vector $v$ with the highest cosine similarity ($S_{cos}$) to $u$. With this choice, the proposed term is able to maximize the similarity between $u$ and the selected vector $v$, while approximating the $id$ of this vector to 1, and the other $id$ to 0 (Eq.~\ref{eq:bonafide}).

\begin{equation}
    Ver_{term} = S_{cos}(v_1, u) > S_{cos}(v_2, u)
\end{equation}

\begin{equation}
\label{eq:bonafide}
L_{KD_1} =
  \begin{cases}
    (1-id_1)^2 + (id_2)^2 - S_{cos}(v_1, u)  & \quad \text{if } Ver_{term}\\
    (1-id_2)^2 + (id_1)^2 - S_{cos}(v_2, u)  & \quad \text{otherwise }
  \end{cases}
\end{equation}

\begin{equation}
    \label{eq:attack}
    L_{KD_2} = [S_{cos}(u_1, u_2)-S_{cos}(v_1, v_2)]^2
\end{equation}

\begin{equation}
    L_{KD} = yL_{KD_1}+(1-y)L_{KD_2}
\end{equation}

Both losses are incorporated in a single equation as follows: 

\begin{equation}
\label{eq:final}
    Loss = L_{BCE} + L_{KD}
\end{equation}

\section{Databases}
This work builds on top of a proposal by Neto et al.~\cite{neto2022orthomad}, hence, we use the same datasets to train and test our methodology:

\begin{enumerate}
    
    \item FRLL: The Face Research London Lab dataset~\cite{debruine_jones_2021} was used to produce the FRLL-Morphs dataset~\cite{sarkar2020vulnerability}, which is frequently used to test morphing attack detection methods. Five different morphing techniques are used in the dataset, including Style-GAN2~\cite{karras2020training,venkatesh2020can}, WebMorph~\cite{debruine2018debruine}, AMSL~\cite{neubert2018extended}, FaceMorpher~\cite{quek2019facemorpher}, and OpenCV~\cite{Mallick_2016}. Each of the five methods uses 204 genuine samples and more than one thousand morphed faces made from high-resolution frontal faces. We used this database only for evaluation purposes because it lacks distinct train, validation or test sets.

    \item SMDD: The Synthetic Morphing Attack Detection Development (SMDD)~\cite{damer2022privacy}, is a novel dataset that uses synthetic images to create a dataset of morph and \textit{bonafide} samples. It initially generated 500k images of faces using a random Gaussian noise vector sampled from a normal distribution using the official open-source version of StyleGan2-ADA~\cite{karras2020training}. Leveraging the quality estimation method known as CR-FIQA~\cite{boutros2021cr}, 50k of these photos were chosen for analysis because of their high quality, and 25k of them were determined to be the \textit{bonafide} samples. The attack photos were paired with five other attack images at random, and 5k of them were chosen as key morphing images. Next, using the OpenCV~\cite{Mallick_2016} method, they were morphed, yielding 15k attack samples. The original 25k images that were used to generate the morphs are also publicly available. This dataset was divided in test and validation sets, on a proportion of 85-15\%.
    
\end{enumerate}

\begin{table*}[ht!]
 \caption{Results comparison with four models published in the literature. All the models were trained on the SMDD dataset, and evaluated on the dataset specified on the left column of the table. All the results are in percentage (\%) and the best are in bold. }
\label{tab:res-rmfd2}
\centering
\resizebox{0.6\textwidth}{!}{
\begin{tabular}{|c|c|c|c|c|}
\hline
& &  & \multicolumn{2}{c|}{BPCER @ APCER =} \\
\cline{1-5}
Test& Model & EER &  1\% &  20\% \\ 
\cline{1-5}
FRLL-Style-GAN2 & \makecell{Inception\\PW-MAD\\MixFacenet\\OrthoMAD\\\textbf{IDistill (Ours)}}& \makecell{11.37\\16.64 \\8.99 \\6.54\\\textbf{1.96}} & \makecell{72.06\\80.39 \\42.16 \\13.74\\\textbf{8.51}}  &  \makecell{6.86\\13.24 \\4.41 \\3.76\\\textbf{0.08}}   \\
 \cline{1-5}
FRLL-WebMorph & \makecell{Inception\\PW-MAD\\MixFacenet\\OrthoMAD\\\textbf{IDistill (Ours)}}& \makecell{9.86\\16.65 \\12.35 \\15.23\\\textbf{4.01}} &   \makecell{53.92\\80.39 \\80.39 \\70.92\\\textbf{14.41}}  &  \makecell{2.94\\13.24 \\7.84 \\9.50\\\textbf{0.33}} \\
 \cline{1-5}
FRLL-OpenCV&\makecell{Inception\\PW-MAD\\MixFacenet\\OrthoMAD\\\textbf{IDistill (Ours)}} & \makecell{5.38\\2.42 \\4.39 \\\textbf{0.73}\\2.46} & \makecell{38.73\\22.06 \\26.47 \\\textbf{0.73}\\6.14} &  \makecell{0.98\\0.49 \\1.47 \\0.32\\\textbf{0.16}} \\
 \cline{1-5}
 FRLL-AMSL& \makecell{Inception\\PW-MAD\\MixFacenet\\OrthoMAD\\\textbf{IDistill (Ours)}} & \makecell{10.79\\15.18 \\15.18 \\14.80\\\textbf{4.00}} & \makecell{72.06\\96.57 \\49.51 \\65.05\\\textbf{21.10}} & \makecell{4.90\\5.88 \\11.76 \\10.89\\\textbf{2.85}}\\
\cline{1-5}
 FRLL-FaceMorpher  & \makecell{Inception\\PW-MAD\\MixFacenet\\OrthoMAD\\\textbf{IDistill (Ours)}} & \makecell{3.17\\ 2.20\\3.87 \\\textbf{0.98}\\2.05} & \makecell{30.39\\ 26.47\\23.53 \\\textbf{2.37}\\4.26} & \makecell{0.49\\ \textbf{0.00}\\0.49 \\0.08\\0.16} \\
\cline{1-5}

\end{tabular}
}
\end{table*}

\section{Experimental setup}

 The autoencoder was trained for $300$ epochs, with a learning rate of $1\times10^{-4}$, a batch size of $32$, and Adam~\cite{kingma2014adam} was used as the optimisation algorithm to minimize the MSE loss. It trained exclusively on \textit{bonafide} samples.

The classifier was trained with the joint loss (Eq.~\ref{eq:final}) utilizing a learning rate of $1\times10^{-4}$, a batch size of 16 and was optimised with Adam. Furthermore, to align with the autoencoder, both $v_1$ and $v_2$ are 128-d. The training utilized the synthetic dataset SMDD, which allowed for this regularization term to utilise the original samples that originated the morphing samples. 

To evaluate the performance of the morphing detection, we evaluated our algorithm using different metrics, commonly used in the literature: the Attack Presentation Classification Error Rate (APCER) (i.e., morphing attacks classified as \textit{bonafide}); and the \textit{Bonafide} Presentation Classification Error Rate (BPCER) (i.e., the \textit{bonafide} samples that are classified as morphing attacks). We evaluated these metrics at two different fixed APCER values (1.0\% and 20.0\%). The equal error rate (EER), which is the BPCER and APCER at the decision thresholds where they are the same, was also evaluated.


\section{Results and Discussion}

The literature on face morphing attack detection is large, however, is also disperse. In other words, the datasets used for benchmarking and training are not always the same, and as such, direct comparisons are not trivial. The combination of FRLL and SMDD has been found in at least two different documents in the literature. The first~\cite{damer2022privacy} introduces the SMDD datasets and evaluate three different methods from the literature: Inception~\cite{szegedy2016rethinking}, PW-MAD~\cite{damer2021pw} and MixFacenet~\cite{boutros2021mixfacenets}. Their results vary and there is not one that beats the others cosistently across the different FRLL morphing methods. Afterwards, OrthoMAD was also evaluated using the exact same protocol~\cite{neto2022orthomad} and achived state-of-the-art results on three out of the five morphing approaches.

Since we follow the protocol introduced by Damer~\textit{et al.}~\cite{damer2022privacy}, the comparison between our method and the ones in the literature focuses on the above mentioned approaches. The results of our method, IDistill, are displayed in Table~\ref{tab:res-rmfd2}. As seen, IDistill has been able to surpass MixFacenet and Inception in all the test databases, and PW-MAD in four out of five databases. OrthoMAD has better results on two databases. A careful analysis of the results highlights an important notion that IDistill is fairly more consistent, as such, the improvements on the databases where it surpasses the literature are much wider than the loss in the performance on the two other databases. Looking at the most extreme examples, in FRLL-OpenCV the EER of our architecture is only 1.73 percentual points larger than the value obtained by OrthoMAD, while IDistill decreases the EER in 11.22 percentual points when tested in the FRLL-WebMorph dataset, which constitutes a much more relevant difference in performance. While looking beyond EER it is also possible to see a wide improvement on the BPCER@APCER at both 1\% and 20\%. Moreover, on FRLL-OpenCV the higher EER of IDistill is mitigated by a lower BPCER@APCER = 20\%. 

When compared to OrthoMAD, our method presents an architecture with the same computation cost on inference, but significantly more interpretable, since OrthoMAD does not guarantee that the information yield by both vectors is related to identity. We are capable of identifying not only attacks, but justify utilising the information of which vectors contain the identity, and which do not. Due to the approximation between the autoencoder latent space and the IDistill latent space, it might also be possible to reconstruct parts of the identity utilising the decoder. While not used in this study, the information regarding the intensity of the vectors extracted by the morphing classifier, $v_1$ and $v_2$ might also allow to infer the morphing percentage associated with each fused identity, which might be useful in future works.

\section{Conclusion}
In this document we have presented a novel method for face morphing attack detection that is interpretable, compact and performs at the state-of-the-art level. The proposed IDistill method was trained utilising a two step scheme based on the training of an autoencoder to reconstruct \textit{bonafide} images, and a distillation step integrated on the standard training of a morphing classifier, utilizing the encoder as teacher and the first part of the classifier as student. 

While we relaxed the orthogonality constraint from previous methods, we devised a more consistent and reliable solution to ensure that the identity information is, in fact, separated in two individual vectors. Moreover, we dismiss any concatenation of these vectors, ensuring an interpretable analysis of the scores produced by each and their contribution to the final prediction. As future work on the interpretability capabilities of this study, it would be interesting to explore the reconstruction capabilities utilising the identity vectors and the decoder model from the autoencoder architecture. Another possible direction is to verify whether the intensities of the vectors extracted by the morphing classifier allow to quantify the morphing percentage of the identities that were fused to generate each attack sample.

Overall, IDistill surpasses the performance of the previous methods published in the literature, while ensuring the advantages previously mentioned. In some scenarios the performance is drastically better. There is much work to be done on the topic of face morphing attack detection, nonetheless, IDistill is a step forward towards the integration of interpretable approaches that are competitive with fully black-box systems.


%

\ifCLASSOPTIONcompsoc
  \section*{Acknowledgments}
\else
  \section*{Acknowledgment}
\fi

This work is co-financed by Component 5 - Capitalization and Business Innovation, integrated in the Resilience Dimension of the Recovery and Resilience Plan within the scope of the Recovery and Resilience Mechanism (MRR) of the European Union (EU), framed in the Next Generation EU, for the period 2021 - 2026, within project NewSpacePortugal, with reference 11. It was also financed by National Funds through the Portuguese funding agency, FCT - Fundação para a Ciência e a Tecnologia within the PhD grants ``2020.06434.BD'' and ``2021.06872.BD''. The research work has been also funded by the German Federal Ministry of Education and Research and the Hessen State Ministry for Higher Education, Research and the Arts within their joint support of the National Research Center for Applied Cybersecurity ATHENE.

\ifCLASSOPTIONcaptionsoff
  \newpage
\fi



\bibliographystyle{IEEEtran}
\bibliography{ref}
%

\end{document}